\documentclass[10pt,twocolumn,letterpaper]{article}

\usepackage{cvpr}
\usepackage{times}
\usepackage{graphicx}
\usepackage{amsmath,amssymb}
\usepackage{booktabs}
\usepackage{multirow}
\usepackage{tabularx}
\usepackage{microtype}
\usepackage{xcolor}
\usepackage{colortbl}
\usepackage{enumitem}
\usepackage{placeins}
\usepackage[pagebackref,breaklinks,colorlinks,allcolors=cvprblue]{hyperref}
\definecolor{cvprblue}{rgb}{0.21,0.49,0.74}
\definecolor{posgreen}{RGB}{0,145,85}
\definecolor{neggray}{RGB}{135,135,135}
\definecolor{methodshade}{RGB}{235,242,248}

\def\confName{CVPR}
\def\confYear{2027}
\newcommand{\method}{SRDP}
\newcommand{\cellvitpp}{\mbox{CellViT++}}
\newcommand{\pos}[1]{\textcolor{posgreen}{#1}}
\newcommand{\decrease}[1]{\textcolor{neggray}{#1}}

\title{Unlocking Fine-Grained Cell Representations via Soft-Interior Rim Dense-Grid Pooling in Pathology Foundation Models}

\author{%
Wenwen Lou\textsuperscript{1,*}, Mingyi He\textsuperscript{2,*},
Xinyi Guo\textsuperscript{3}, Yining Li\textsuperscript{4},
Weiming Chen\textsuperscript{1}\\
Jiawen Li\textsuperscript{1}, Lianghui Zhu\textsuperscript{1},
Zhiwei Yang\textsuperscript{5,$\dagger$},
Xitong Ling\textsuperscript{1,$\dagger$}\\
\textsuperscript{1}Tsinghua University,
\textsuperscript{2}Beijing University of Chemical Technology,
\textsuperscript{3}South China Normal University\\
\textsuperscript{4}Nankai University,
\textsuperscript{5}Peking Union Medical College
}

\hypersetup{
  pdftitle={Unlocking Fine-Grained Cell Representations via Soft-Interior Rim Dense-Grid Pooling in Pathology Foundation Models},
  pdfauthor={Wenwen Lou, Mingyi He, Xinyi Guo, Yining Li, Weiming Chen, Jiawen Li, Lianghui Zhu, Zhiwei Yang, Xitong Ling},
  pdfsubject={Fine-grained cell representations from frozen pathology foundation models}
}

\begin{document}
\maketitle
\begin{NoHyper}
\begingroup
\renewcommand{\thefootnote}{}
\footnotetext{\textsuperscript{*}Contributed equally. \textsuperscript{$\dagger$}Corresponding authors.}
\endgroup
\end{NoHyper}
\setcounter{footnote}{0}
\raggedbottom

\begin{abstract}
Fine-grained cell representations are essential for characterizing cellular heterogeneity and tissue organization in computational pathology. Pathology foundation models (PFMs) provide powerful pretrained features for transferable tissue analysis. HoVer-Net uses task-trained segmentation and typing, whereas \cellvitpp{} averages native-grid tokens assigned to detected nuclei. However, the former requires cell-specific training; the latter ignores fractional nuclear coverage and can mix nuclear content with neighbouring tissue. Therefore, we propose \method{}, short for \textbf{S}oft-interior \textbf{R}im \textbf{D}ense-grid \textbf{P}ooling, a training-free geometry-aware readout. We apply the frozen patch projection at half stride, interleave four offset lattices, map the cell mask to fractional coverage weights, and concatenate whole-cell pooling with two inner boundary pools. Across four datasets and six frozen encoders, SRDP improves mean balanced accuracy by 5.68 percentage points over native token averaging across 24 settings, including 11.22 points on average for CoNIC and Lizard. SRDP converts a coarse frozen PFM lattice into a mask-aligned, fine-grained cell representation without updating the encoder.
\end{abstract}

\section{Introduction}
Nuclear shape, chromatin, cytoplasm, and local context inform cell phenotyping~\cite{mandal2025glossary}. Lizard~\cite{graham2021lizard} and CoNIC~\cite{conic2022} annotate colorectal nuclei and types. PFMs such as UNI~\cite{chen2024uni}, CONCH~\cite{lu2023conch}, and Prov-GigaPath~\cite{xu2024gigapath} provide transferable features. HoVer-Net~\cite{graham2019hovernet} jointly trains nuclear-pixel, distance, and type branches over a convolutional encoder. \cellvitpp{}~\cite{hoerst2026cellvitpp} freezes a pretrained CellViT~\cite{hoerst2024cellvit} segmenter, maps detected nuclei to final-layer tokens, averages embeddings for multi-token nuclei, and learns a lightweight classifier. We ask whether masks enable finer readout from a frozen PFM.

The obstacle is spatial rather than semantic: a nucleus may be smaller than one patch token, cross a token boundary, or share a token with cytoplasm and neighbouring cells. Native mask pooling then supplies few samples and counts a token as cellular even when the mask covers only a small part of its receptive field. Reducing the inference stride exposes a denser frozen ViT grid~\cite{amir2022deep}, while pixel-as-token transformers show that individual pixels can replace fixed image patches~\cite{nguyen2025pixels}. The remaining problem is to align an irregular cell with that grid while preserving both interior and contour morphology.

\begin{figure}[t]
  \centering
\includegraphics[width=\linewidth]{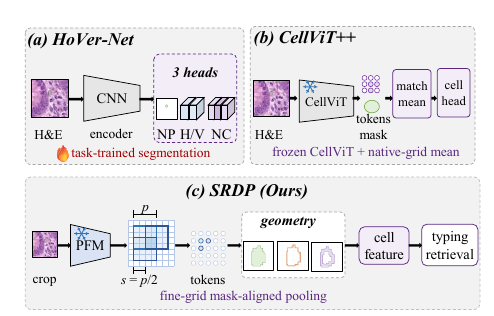}
\caption{\textbf{Cell-level representation readouts.} (a) HoVer-Net~\cite{graham2019hovernet} jointly trains nuclear-pixel, distance, and type branches. (b) \cellvitpp{}~\cite{hoerst2026cellvitpp} freezes CellViT~\cite{hoerst2024cellvit}, averages assigned final-layer tokens per nucleus, and trains a cell head. (c) \method{} pools fractional whole-cell and inner-rim support from four frozen half-stride lattices. Snowflakes mark frozen modules; the flame marks training.}
  \label{fig:teaser}
\end{figure}

Our key observation is that overlapping patches observe the same location through translated neighbourhoods. We introduce \textbf{S}oft-interior \textbf{R}im \textbf{D}ense-grid \textbf{P}ooling (\method{}), which halves the frozen patch projection stride, converts coverage of the cell mask into continuous grid weights, and pools the whole cell together with two inner rims. Concatenating the three branches preserves interior and contour evidence in one fixed descriptor. The frozen backbone and deterministic readout require no representation training. Figure~\ref{fig:teaser} contrasts task-trained HoVer-Net, native-grid \cellvitpp{} readout, and our dense mask-aligned readout; Fig.~\ref{fig:method} details SRDP.

We make three contributions. First, we formulate fine-grained PFM extraction as a geometry-aware readout problem. Second, we unify translated overlap sampling, fractional mask alignment, and multiscale inner-rim pooling in a deterministic descriptor. Third, we evaluate four cell datasets and six frozen encoders through linear probing, few-shot classification, direct retrieval, and targeted ablations. Mean balanced accuracy improves by 5.68 points across 24 dataset and encoder settings, including an average gain of 11.22 points on CoNIC~\cite{conic2022} and Lizard~\cite{graham2021lizard}.

\section{Related Work}
\paragraph{Pathology foundation models.}
Patch-level PFMs preserve local token lattices. KimiaNet~\cite{riasatian2021kimianet} uses weak labels; Ciga et al.~\cite{ciga2022ssl} learn histology contrast; RetCCL~\cite{wang2022retccl} adds cluster guidance; CTransPath~\cite{wang2022ctranspath} combines CNN and Swin encoders. REMEDIS~\cite{azizi2023remedis} transfers across six medical imaging domains. PathoDuet~\cite{hua2023pathoduet} models scale and stain relations, whereas Phikon~\cite{filiot2023phikon} uses iBOT. Kaiko~\cite{aben2024largescale} studies scaling; Phikon-v2~\cite{filiot2024phikonv2}, Hibou~\cite{nechaev2024hibou}, and UNI~\cite{chen2024uni} use DINOv2-style objectives. RudolfV~\cite{dippel2024rudolfv} adds pathologist-guided multi-stain curation. Virchow~\cite{vorontsov2024virchow} and Virchow2~\cite{zimmermann2024virchow2} scale data and magnifications. PLUTO~\cite{juyal2024pluto} is compact and multiscale; GPFM~\cite{ma2024gpfm} distils expert and self knowledge; Atlas~2~\cite{alber2026atlas2} uses 5.5 million slides. PathOrchestra~\cite{yan2025pathorchestra} spans 112 tasks, while Digepath~\cite{zhu2026digepath} targets gastrointestinal pathology.

Whole-slide PFMs combine tile embeddings to model tissue context. HIPT~\cite{chen2022hipt} learns across nested scales. Prov-GigaPath~\cite{xu2024gigapath} couples a tile encoder to a long-context slide encoder. CHIEF~\cite{wang2024chief} uses weak slide and anatomical site supervision; PRISM~\cite{shaikovski2024prism} aligns slides with reports; MADELEINE~\cite{jaume2024madeleine} aligns slides across stains.

Multimodal PFMs extend CLIP-style contrast~\cite{radford2021clip}. PLIP~\cite{huang2023plip}, QuiltNet~\cite{ikezogwo2023quilt}, BiomedCLIP~\cite{zhang2023biomedclip}, CONCH~\cite{lu2023conch}, and MUSK~\cite{xiang2025musk} learn from biomedical image--text pairs; MI-Zero~\cite{lu2023mizero} aggregates slide evidence. TITAN~\cite{ding2025titan} aligns regions and slides with text. TANGLE~\cite{jaume2024tangle} uses transcriptomes; mSTAR~\cite{xu2025mstar} adds reports and genes; Threads~\cite{vaidya2025threads} uses molecular profiles; WILSON~\cite{alfasly2026wilson} links multi-slide cases to reports. Benchmarks cover weak supervision~\cite{neidlinger2024benchmark}, precision oncology~\cite{ma2025pathbench}, and spatial analysis~\cite{zhao2026spapath}. We query frozen lattices at cell scale.

\paragraph{Cell representations.}
StarDist~\cite{schmidt2018stardist} predicts star-convex polygons, Cellpose~\cite{stringer2021cellpose} predicts flows, and HoVer-Net~\cite{graham2019hovernet} separates nuclei with directional distances. PanNuke~\cite{gamper2019pannuke} supplies pan-cancer labels. CellViT~\cite{hoerst2024cellvit} couples a ViT encoder to nucleus segmentation decoders. \cellvitpp{}~\cite{hoerst2026cellvitpp} freezes that segmenter, averages its assigned final-layer tokens per nucleus, and trains a lightweight classifier. Nuc2Vec~\cite{feng2021nuc2vec}, VOLTA~\cite{nakhli2024volta}, and LEMON~\cite{chadoutaud2026lemon} learn cell morphology embeddings; CellDETR~\cite{zhang2026celldetr} uses detection-guided attention.

\paragraph{Dense and fine-grained features.}
ViTs~\cite{dosovitskiy2021vit} encode patch sequences. Masked autoencoders~\cite{he2022mae}, DINO~\cite{caron2021dino}, iBOT~\cite{zhou2022ibot}, DINOv2~\cite{oquab2023dinov2}, and DINOv3~\cite{simeoni2025dinov3} learn transferable features; register tokens reduce map artifacts~\cite{darcet2024registers}. DPT~\cite{ranftl2021dpt} fuses multistage tokens, while OPSNet~\cite{chen2023opsnet} pools spatial CLIP features within masks. Amir et al.~\cite{amir2022deep} reduce inference stride, FeatUp~\cite{fu2024featup} learns upsampling, pixel-as-token models~\cite{nguyen2025pixels} remove patchification, and CMD~\cite{chen2026cmd} uses convolutional masked-diffusion pretraining. We retain the frozen projection and add continuous coverage with inner rims.

\paragraph{Nucleus geometry.}
CoNIC~\cite{conic2022} and Lizard~\cite{graham2021lizard} label colorectal nucleus instances and types; NuCLS~\cite{amgad2021nucls} labels breast cancer nuclei; MoNuSAC~\cite{verma2021monusac} spans four organs and four types. Shape, chromatin, cytoplasm, and context inform cell identity~\cite{mandal2025glossary}, motivating interior and rim supports.

\paragraph{Closest spatial readouts.}
CELLO~\cite{gao2026cello} samples PFM maps on a grid at nuclei and refines cell features using cross-attention with distance decay for transcriptomic prediction. CellPath-Bench~\cite{zhao2026cellpath} samples frozen WSI maps at registered nuclei to measure cell-type decodability and transfer. We improve extraction through translated sampling, fractional mask coverage, and fixed inner rims.

\section{SRDP Method}
\subsection{Problem Setup and Cell-Centred View}
Let $I_i$ be an H\&E image containing a cell with binary mask $M_i$ and optional label $y_i$. A pretrained encoder $f_\theta$ produces descriptor $z_i$ through a fixed spatial readout; a downstream linear probe does not alter its extraction.

\begin{figure*}[t]
  \centering
  \includegraphics[width=\textwidth]{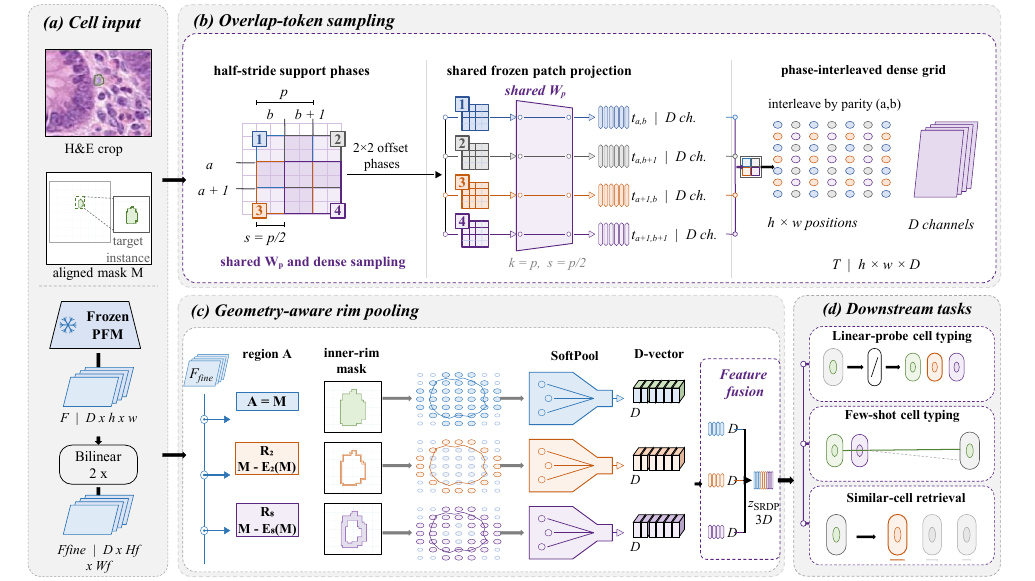}
  \caption{\textbf{The SRDP framework.} (a) A cell-centred crop and aligned mask enter a frozen PFM. (b) SRDP reuses the frozen patch projection at half stride and interleaves four translated lattices into a dense grid. (c) Fractional mask weights pool the whole cell and two inner rims before concatenation. (d) The descriptor supports linear probing, few-shot classification, and cosine retrieval.}
  \label{fig:method}
\end{figure*}

The mask defines the readout coordinates: Fig.~\ref{fig:method} passes a cell-centred crop through the frozen encoder, forms a dense lattice from overlapping projections, and pools fixed mask supports. The same cached descriptor serves fitted probes and retrieval.

Our design gives small cells more samples, weights partial coverage proportionally, and keeps contour morphology distinct from the whole-cell average.

We first convert an arbitrarily located cell into an encoder-compatible view. Let the native square input size be $S$, the patch size be $p$, and $(w_i,h_i)$ the cell bounding-box size. A square window of side
\begin{equation}
L_i=\max\!\left(S,\;p\left\lceil\frac{\max(w_i,h_i)}{p}\right\rceil\right)
\end{equation}
is centred at the bounding box. We zero pad outside the image and resize unusually large, patch-aligned windows to $S\times S$. Bicubic image and nearest-neighbour mask resizing produce the aligned pair $(x_i,M_i)$. The crop retains context, while the readout pools masked locations.

\subsection{Frozen Dense-Grid Extraction}
Following Amir et al.~\cite{amir2022deep}, we treat ViT patch embedding as a convolution with kernel $p$ and set its inference stride to $s=\frac{p}{2}$. Let $W_p$ and $b_p$ denote the frozen patch projection kernel and bias. The token at grid indices $(a,b)$ is
\begin{equation}
t_{a,b}=W_p\,x_i[as:as+p,\,bs:bs+p]+b_p,
\end{equation}
and adjacent receptive fields overlap. For an $S\times S$ input, the grid grows from $\frac{S}{p}$ to $\left\lfloor\frac{S-p}{s}\right\rfloor+1$ locations per axis. Half-stride sampling observes small or boundary-crossing cells through translated neighbourhoods. Absolute positional embeddings are interpolated when required; models with dynamic positional encodings or rotary embeddings retain their native mechanisms. All encoders run without gradients.

Overlap acts before the nonlinear encoder blocks: the frozen projection measures each nucleus at multiple patch phases. Post-encoding interpolation only aligns these measurements with the fractional mask supports; it does not create new patch-phase observations.

We discard non-spatial summary tokens and obtain
\begin{equation}
F=f_\theta(x_i)\in\mathbb{R}^{D\times h\times w},
\end{equation}
where $D$ is the feature dimension and $h,w$ are the spatial grid dimensions. We compute $F_{\mathrm{fine}}=\operatorname{Bilinear}(F,2)$ once and share it across all readout branches.

\subsection{Soft-Interior Pooling}
A hard centre rule assigns an entire location to a cell. For binary support $A$, we instead compute
\begin{equation}
w_A=\operatorname{AdaptiveAvgPool}(A;H_f,W_f),
\end{equation}
where $H_f$ and $W_f$ are the height and width of $F_{\mathrm{fine}}$, $u$ indexes a grid location, and each $w_A(u)\in[0,1]$ records fractional coverage. Let $Z_A=\sum_u w_A(u)$ denote the total covered support. The area-weighted descriptor is
\begin{align}
\mu(A,F_{\mathrm{fine}}) &=
\begin{cases}
Z_A^{-1}\sum_u w_A(u)F_{\mathrm{fine}}(u),
& Z_A\ge\epsilon,\\[4pt]
\dfrac{1}{H_fW_f}\sum_u F_{\mathrm{fine}}(u),
& Z_A<\epsilon,
\end{cases}
\raisetag{27pt}
\end{align}
where $\epsilon=10^{-6}$ is a numerical threshold. Boundary locations contribute in proportion to cell coverage; support normalization handles cell size, and empty support returns the spatial feature mean.

The mask supplies spatial assignment, while appearance remains encoded by the frozen features.

\subsection{Rim Pooling}
Nuclear contours carry type-specific morphology, including elongation, roundness, and granulocyte lobulation~\cite{mandal2025glossary}; the whole-cell pool retains complementary chromatin and perinuclear appearance.

Let morphological erosion of pixel radius $r$ be
\begin{equation}
E_r(M)=\operatorname{MinPool}_{2r+1}(M).
\label{eq:erosion}
\end{equation}
The corresponding inner boundary band is
\begin{equation}
R_r(M)=M\setminus E_r(M).
\label{eq:rim}
\end{equation}
using same-size padding; we implement the operation by negated max pooling. The inner band $R_r$ stays within the annotated cell. We concatenate one soft whole-cell branch and two rim scales:
\begin{align}
z_{\mathrm{soft}}&=\mu(M_i,F_{\mathrm{fine}}),\\
z_{2}&=\mu(R_2(M_i),F_{\mathrm{fine}}),\\
z_{8}&=\mu(R_8(M_i),F_{\mathrm{fine}}),\\
z_{\mathrm{SRDP}}&=[z_{\mathrm{soft}};z_2;z_8]\in\mathbb{R}^{3D}.
\label{eq:srdp}
\end{align}
The $r=2$ branch emphasizes immediate contour irregularity, while $r=8$ retains a broader contour band. We fix the radii across every dataset and encoder. Concatenation preserves branch identity in a deterministic $3D$ descriptor from one frozen pass, one interpolation, and three adaptive pools.

\section{Experiments}
We evaluate SRDP across encoders, label budgets, and downstream tasks, then isolate its spatial components and extraction cost. We use matched cells and frozen features in every comparison.

\subsection{Data and Evaluation Protocol}
We evaluate CoNIC~\cite{conic2022}, Lizard~\cite{graham2021lizard}, NuCLS~\cite{amgad2021nucls}, and MoNuSAC~\cite{verma2021monusac} with DINOv3-B and DINOv3-L~\cite{simeoni2025dinov3}, CONCH~\cite{lu2023conch}, CONCH v1.5~\cite{ding2025titan}, Prov-GigaPath~\cite{xu2024gigapath}, and UNI~\cite{chen2024uni}. CoNIC and Lizard use fixed image-level partitions, NuCLS follows its official patient-disjoint fold, and MoNuSAC follows its official image split. Every cell from one image remains in the same partition, and all encoders and readouts use identical assignments. Detailed split composition, label filtering, patient grouping, and feature coverage appear in supp. Tables~\ref{tab:splits} and~\ref{tab:coverage}.

\textbf{Token mean}, used by \cellvitpp{}~\cite{hoerst2026cellvitpp}, assigns each cell to the native PFM tokens intersecting its mask and averages their embeddings equally. A barely overlapping token therefore contributes as much as one centred inside the cell. \textbf{Slide pooling} encodes the image at once when possible; larger images are split into disjoint windows matching the encoder input and padded only at the borders. It averages native tokens intersecting each cell mask and combines their sums and counts across window boundaries. Because cells are not recentered in separate crops, the original patch alignment is preserved. \textbf{SRDP} uses Eq.~\ref{eq:srdp}. Within each dataset and encoder pair, all readouts share ordered cells and split labels. We fit a standardizer using training data only, followed by class-balanced multinomial logistic regression with an $\ell_2$ penalty. Its inverse regularization strength is $C=1$, and we optimize it with limited-memory BFGS (L-BFGS). We report balanced accuracy (BAcc), macro-F1 (mF1), and accuracy (Acc).

We select the global rim pair on grouped training-only validation and use $(2,8)$ throughout. The selection pools four datasets and two encoders while preserving image or patient groups, so one morphology scale is fixed before test evaluation. The candidate radii, validation scores, and selection protocol are reported in supp. Sec.~\ref{sec:supp:radius} and supp. Table~\ref{tab:radius-selection}.

\begin{table*}[t]
\centering
\caption{\textbf{Linear-probe results (\%).} Token mean, slide pooling, and SRDP use the same ordered cells and split. Each block reports the SRDP change $\Delta$ over token mean; bold marks the best of the three readouts.}
\label{tab:main}
\footnotesize
\setlength{\tabcolsep}{2.6pt}
\renewcommand{\arraystretch}{0.96}
\resizebox{\textwidth}{!}{%
\begin{tabular}{llrrrrrrrrrrrr}
\toprule
\multirow{2}{*}{Encoder} & \multirow{2}{*}{Readout} & \multicolumn{3}{c}{CoNIC~\cite{conic2022}} & \multicolumn{3}{c}{Lizard~\cite{graham2021lizard}} & \multicolumn{3}{c}{NuCLS~\cite{amgad2021nucls}} & \multicolumn{3}{c}{MoNuSAC~\cite{verma2021monusac}} \\
\cmidrule(lr){3-5}\cmidrule(lr){6-8}\cmidrule(lr){9-11}\cmidrule(lr){12-14}
& & BAcc & mF1 & Acc & BAcc & mF1 & Acc & BAcc & mF1 & Acc & BAcc & mF1 & Acc \\
\midrule
\multirow{4}{*}{DINOv3-B~\cite{simeoni2025dinov3}} & Token mean & 60.60 & 44.89 & 65.47 & 58.07 & 45.19 & 67.95 & \textbf{38.86} & 37.43 & 58.73 & 50.09 & 49.93 & 92.51 \\
 & Slide & 62.12 & 46.91 & 67.41 & 59.28 & 47.60 & 70.21 & 37.02 & 34.18 & 57.26 & 50.50 & 50.15 & 91.02 \\
 & \method{} & \textbf{71.96} & \textbf{57.43} & \textbf{74.99} & \textbf{70.50} & \textbf{58.86} & \textbf{77.36} & 38.77 & \textbf{38.07} & \textbf{59.57} & \textbf{50.66} & \textbf{50.69} & \textbf{93.37} \\
 & $\Delta$ & \pos{+11.37} & \pos{+12.55} & \pos{+9.52} & \pos{+12.43} & \pos{+13.68} & \pos{+9.42} & \decrease{-0.09} & \pos{+0.63} & \pos{+0.84} & \pos{+0.57} & \pos{+0.76} & \pos{+0.86} \\
\midrule
\multirow{4}{*}{DINOv3-L~\cite{simeoni2025dinov3}} & Token mean & 60.71 & 45.43 & 66.45 & 57.72 & 45.20 & 68.19 & \textbf{40.11} & \textbf{38.89} & 60.68 & 50.53 & 50.42 & 92.80 \\
 & Slide & 62.50 & 47.22 & 67.95 & 60.38 & 48.08 & 70.51 & 37.39 & 36.34 & 58.22 & \textbf{51.72} & \textbf{51.52} & 92.36 \\
 & \method{} & \textbf{71.66} & \textbf{57.41} & \textbf{75.62} & \textbf{70.71} & \textbf{58.67} & \textbf{77.55} & 38.37 & 38.56 & \textbf{60.70} & 50.53 & 50.67 & \textbf{93.30} \\
 & $\Delta$ & \pos{+10.95} & \pos{+11.97} & \pos{+9.17} & \pos{+13.00} & \pos{+13.46} & \pos{+9.36} & \decrease{-1.74} & \decrease{-0.33} & \pos{+0.03} & \decrease{-0.01} & \pos{+0.25} & \pos{+0.50} \\
\midrule
\multirow{4}{*}{CONCH~\cite{lu2023conch}} & Token mean & 70.69 & 54.33 & 75.00 & 70.50 & 55.59 & 77.16 & 40.24 & 38.43 & 62.31 & \textbf{50.05} & 50.17 & \textbf{94.43} \\
 & Slide & 72.11 & 55.98 & 75.88 & 71.26 & 56.91 & 77.89 & 40.17 & 37.62 & 60.90 & 49.78 & 49.83 & 94.39 \\
 & \method{} & \textbf{82.46} & \textbf{69.23} & \textbf{84.78} & \textbf{82.82} & \textbf{71.64} & \textbf{86.62} & \textbf{41.30} & \textbf{40.46} & \textbf{63.00} & 50.05 & \textbf{50.19} & 94.31 \\
 & $\Delta$ & \pos{+11.77} & \pos{+14.91} & \pos{+9.78} & \pos{+12.33} & \pos{+16.05} & \pos{+9.47} & \pos{+1.07} & \pos{+2.03} & \pos{+0.69} & \decrease{-0.01} & \pos{+0.02} & \decrease{-0.12} \\
\midrule
\multirow{4}{*}{CONCH v1.5~\cite{ding2025titan}} & Token mean & 68.70 & 52.84 & 73.83 & 69.11 & 55.17 & 76.83 & 37.61 & 37.00 & 60.84 & 49.71 & 49.72 & 94.61 \\
 & Slide & 71.26 & 55.55 & 75.97 & 70.63 & 56.20 & 77.36 & \textbf{40.06} & \textbf{38.55} & 61.81 & 49.65 & 49.62 & 93.73 \\
 & \method{} & \textbf{81.11} & \textbf{68.17} & \textbf{84.69} & \textbf{80.05} & \textbf{71.01} & \textbf{86.68} & 38.37 & 38.16 & \textbf{62.71} & \textbf{49.87} & \textbf{49.97} & \textbf{94.64} \\
 & $\Delta$ & \pos{+12.40} & \pos{+15.34} & \pos{+10.86} & \pos{+10.94} & \pos{+15.84} & \pos{+9.85} & \pos{+0.77} & \pos{+1.16} & \pos{+1.86} & \pos{+0.16} & \pos{+0.25} & \pos{+0.02} \\
\midrule
\multirow{4}{*}{GigaPath~\cite{xu2024gigapath}} & Token mean & 73.56 & 58.48 & 78.40 & 71.87 & 58.97 & 79.57 & \textbf{38.95} & \textbf{38.75} & \textbf{62.16} & 49.69 & 49.59 & 94.11 \\
 & Slide & 74.04 & 58.92 & 78.40 & 73.10 & 59.94 & 80.09 & 38.04 & 37.05 & 58.85 & 49.39 & 49.35 & 93.63 \\
 & \method{} & \textbf{83.69} & \textbf{73.26} & \textbf{87.26} & \textbf{81.46} & \textbf{73.86} & \textbf{88.22} & 36.70 & 37.62 & 59.75 & \textbf{50.28} & \textbf{50.22} & \textbf{94.68} \\
 & $\Delta$ & \pos{+10.13} & \pos{+14.78} & \pos{+8.86} & \pos{+9.60} & \pos{+14.89} & \pos{+8.65} & \decrease{-2.25} & \decrease{-1.13} & \decrease{-2.41} & \pos{+0.59} & \pos{+0.63} & \pos{+0.57} \\
\midrule
\multirow{4}{*}{UNI~\cite{chen2024uni}} & Token mean & 72.55 & 56.50 & 77.18 & 73.09 & 58.87 & 79.55 & 49.41 & 39.02 & 63.28 & \textbf{49.54} & \textbf{49.54} & \textbf{94.51} \\
 & Slide & 73.66 & 58.19 & 78.23 & 74.00 & 59.48 & 79.71 & 51.36 & 40.65 & 63.78 & 49.35 & 49.37 & 93.75 \\
 & \method{} & \textbf{83.01} & \textbf{70.33} & \textbf{86.01} & \textbf{82.39} & \textbf{72.05} & \textbf{87.21} & \textbf{52.05} & \textbf{42.28} & \textbf{65.19} & 49.41 & 49.45 & 94.09 \\
 & $\Delta$ & \pos{+10.46} & \pos{+13.84} & \pos{+8.83} & \pos{+9.30} & \pos{+13.18} & \pos{+7.66} & \pos{+2.63} & \pos{+3.25} & \pos{+1.91} & \decrease{-0.13} & \decrease{-0.10} & \decrease{-0.42} \\
\bottomrule

\end{tabular}%
}
\end{table*}

\subsection{Main Results}
Table~\ref{tab:main} shows the clearest gains on CoNIC~\cite{conic2022} and Lizard~\cite{graham2021lizard}. Averaged over both datasets and all six encoders, we improve BAcc by $11.22$ points and mF1 by $14.21$ points over token mean. We also gain $4.97$ BAcc points over slide pooling on average across all 24 settings, and $9.79$ points on the two colorectal datasets. We observe this consistency across general-purpose and pathology-specific encoders, identifying the readout as the shared source of improvement. Translated sampling and fractional coverage prevent local nuclear cues from collapsing into a few mixed tokens, raising mean BAcc by $5.68$ points over token mean across all settings. NuCLS~\cite{amgad2021nucls} varies by encoder, whereas the four-label MoNuSAC~\cite{verma2021monusac} evaluation remains nearly tied across readouts.

On both colorectal datasets, mF1 gains exceed BAcc gains for every encoder, indicating sharper class-specific boundaries rather than an advantage confined to common classes. The encoder-dependent response on NuCLS, together with the near ties on MoNuSAC, shows that dense geometric readout is most effective when the frozen lattice retains label-relevant morphology at cell scale.

We also verify that the improvement is not tied to one metric. Across CoNIC and Lizard, we improve all 36 comparisons of dataset, encoder, and metric in Table~\ref{tab:main}; Acc rises by $9.28$ points on average alongside the larger BAcc and mF1 gains. On NuCLS, we obtain the strongest SRDP result with UNI at $52.05$ BAcc, a $2.63$-point gain over token mean, while other encoders change less or decline. The backbone-dependent variation indicates that the readout is most effective when the frozen features already encode cell morphology.

\subsection{Few-Shot Cell Classification}
\begin{figure}[!t]
  \centering
  \includegraphics[width=\linewidth]{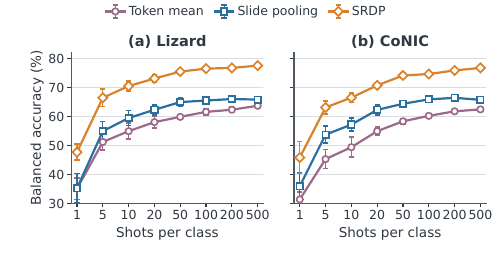}
  \caption{\textbf{Few-shot classification with Prov-GigaPath~\cite{xu2024gigapath}.} (a) Lizard~\cite{graham2021lizard}. (b) CoNIC~\cite{conic2022}. In both datasets, SRDP remains above token mean and slide pooling from one to 500 shots per class. Curves show mean BAcc and standard deviation over ten matched class-balanced support draws.}
  \label{fig:fewshot}
\end{figure}

We test whether the representation remains useful when labels are scarce: on Lizard~\cite{graham2021lizard} and CoNIC~\cite{conic2022} with Prov-GigaPath~\cite{xu2024gigapath}, we sample the same class-balanced support set for all readouts at each budget from one to 500 examples per class. Ten independent support draws share a fixed test set; only the probe is refitted. We use this protocol to isolate the descriptor from differences in labelled cells and expose variation across draws. Figure~\ref{fig:fewshot} reports the mean and standard deviation.

SRDP ranks first at every shot count in Fig.~\ref{fig:fewshot}. Its smallest lead over the stronger baseline is $10.54$ BAcc points on Lizard and $8.47$ on CoNIC. Under the same encoder, this consistent ordering isolates the readout: overlap sampling supplies more localized observations, while aligned interior and rim pools expose morphology before the probe sees labels. At 500 shots per class, SRDP still leads by $11.73$ points on Lizard and $11.05$ on CoNIC. More labels improve every probe but do not restore the cell detail lost by coarse token pooling.

We use the unchanged ordering across label budgets to separate representation quality from probe supervision and identify mask-aligned descriptor geometry as the source of improvement.

\subsection{Training-Free Cell Retrieval}
We next remove the fitted classifier entirely: for each dataset and encoder, training cells form the gallery and test cells are queries, and ranking uses cosine similarity between cached descriptors. We use the same gallery, query set, and backbone features for every readout. Query-weighted R@1 can be dominated by common classes, so Table~\ref{tab:retrieval} also reports macro R@1 (mR@1), the unweighted mean of class-specific hit rates. We then average both metrics uniformly across six encoders.

\begin{table}[!h]
  \centering
  \caption{\textbf{Training-free cell retrieval (\%).} Recall at rank 1 (R@1) and its unweighted class mean (mR@1) are averaged over six frozen encoders.}
  \label{tab:retrieval}
  \footnotesize
  \setlength{\tabcolsep}{2.2pt}
  \renewcommand{\arraystretch}{1.02}
  \begin{tabularx}{\linewidth}{@{}ll*{4}{>{\centering\arraybackslash}X}@{}}
    \toprule
    Readout & Metric & \shortstack{CoNIC\\\cite{conic2022}} & \shortstack{Lizard\\\cite{graham2021lizard}} & \shortstack{NuCLS\\\cite{amgad2021nucls}} & \shortstack{MoNuSAC\\\cite{verma2021monusac}} \\
    \midrule
    \multirow{2}{*}{Token mean} & R@1 & 73.26 & 68.93 & \textbf{59.02} & \textbf{93.62} \\
 & mR@1 & 54.10 & 44.38 & 32.84 & 50.72 \\
\multirow{2}{*}{Slide pooling} & R@1 & 73.30 & 70.69 & 57.41 & 93.30 \\
 & mR@1 & 54.88 & 48.26 & 32.01 & \textbf{51.40} \\
\midrule
\rowcolor{methodshade} & R@1 & \textbf{78.38} & \textbf{74.87} & 57.69 & 93.07 \\
\rowcolor{methodshade}\multirow{-2}{*}{\method{} (ours)} & mR@1 & \textbf{64.02} & \textbf{57.10} & \textbf{33.18} & 49.85 \\
\multirow{2}{*}{$\Delta$ vs. token} & R@1 & \pos{+5.12} & \pos{+5.94} & \decrease{-1.33} & \decrease{-0.56} \\
 & mR@1 & \pos{+9.92} & \pos{+12.72} & \pos{+0.33} & \decrease{-0.87} \\
\bottomrule

  \end{tabularx}
\end{table}

We observe improvements in both retrieval metrics on CoNIC~\cite{conic2022} and Lizard~\cite{graham2021lizard}. Averaged over six encoders, CoNIC gains $5.12$ R@1 points and $9.92$ mR@1 points; Lizard gains $5.94$ R@1 points and $12.72$ mR@1 points. The larger macro gains show that less frequent classes benefit most. On NuCLS~\cite{amgad2021nucls}, R@1 changes by $-1.33$ points and mR@1 by $+0.33$ points, while MoNuSAC~\cite{verma2021monusac} improves neither metric. Results at ranks 5 and 10 are reported in supp. Table~\ref{tab:macro-retrieval}.

Because retrieval uses no fitted decision layer, these results test the geometry of the descriptor space directly. Our readout brings same-class colorectal cells closer under cosine similarity even when the query belongs to a less frequent phenotype, which explains why mR@1 rises more than query-weighted R@1. On NuCLS and MoNuSAC, the smaller retrieval changes track the corresponding linear-probe results, connecting the benefit of the readout to how strongly each label is expressed by interior and contour morphology.

\begin{figure}[t]
  \centering
  \includegraphics[width=\linewidth]{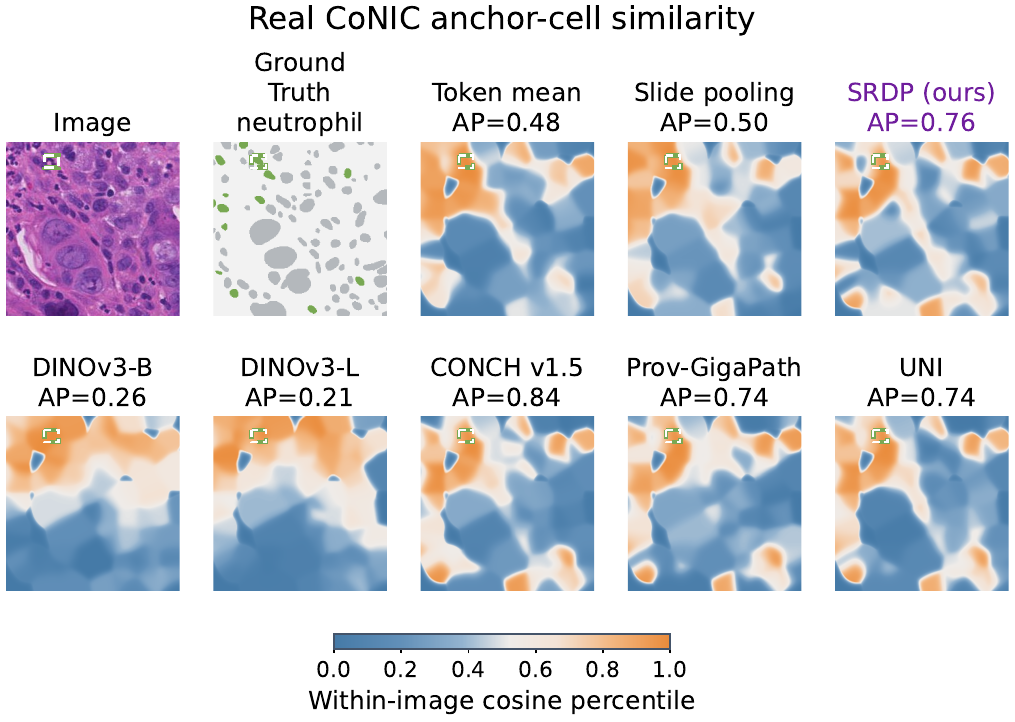}
\caption{\textbf{Similarity to an anchor cell in a real CoNIC~\cite{conic2022} field.} The top row locates the selected neutrophil anchor and compares token mean, slide pooling, and SRDP with CONCH~\cite{lu2023conch}. The bottom row applies SRDP to the other frozen encoders. All heatmaps share one percentile scale of cosine similarity; interpolation is used only for display.}
  \label{fig:anchor_similarity}
\end{figure}

\begin{figure*}[t]
\centering
\includegraphics[width=\textwidth]{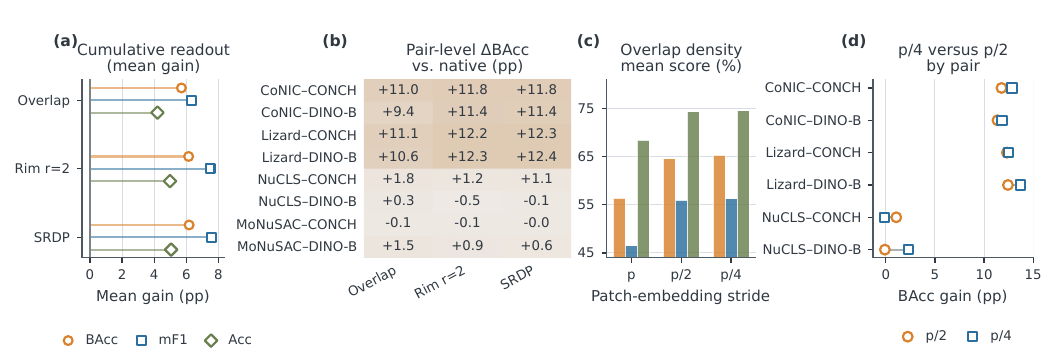}
\caption{\textbf{Cumulative readout and sampling density ablations.} (a,b) Mean and pair-level BAcc gains show that overlap supplies most improvement. (c,d) Absolute and pairwise results saturate beyond the half-stride setting. DINO-B denotes DINOv3-B~\cite{simeoni2025dinov3}, and pp denotes percentage points.}
\label{fig:ablation}
\end{figure*}

\paragraph{Similarity to an anchor cell.}
We visualize the descriptor space for a real CoNIC~\cite{conic2022} neutrophil in Fig.~\ref{fig:anchor_similarity}. The field contains 94 annotated cells, including the anchor and 13 other neutrophils, and average precision (AP) measures how highly those same-class cells rank among all cells in the field. With CONCH~\cite{lu2023conch}, we raise AP from $0.481$ for token mean and $0.503$ for slide pooling to $0.756$. In the heatmap, we observe that native and slide-level pooling spread high similarity across mixed neighbouring nuclei, whereas our readout concentrates high responses on cells sharing the anchor phenotype. Our overlapping projections give a boundary-crossing nucleus several localized observations, and the aligned whole-cell and rim pools preserve both its interior appearance and lobulated contour. Their concatenation therefore groups morphologically matched neutrophils before any classifier is fitted. Our readout reaches AP values of $0.842$, $0.740$, and $0.740$ with CONCH v1.5~\cite{ding2025titan}, Prov-GigaPath~\cite{xu2024gigapath}, and UNI~\cite{chen2024uni}, respectively, showing that the proposed geometry supports more selective cell matching with strong frozen pathology features.

\subsection{Ablation Studies}
We compare specialized cell models, add spatial supports cumulatively, vary grid density, perturb mask geometry, and profile extraction. We use CoNIC~\cite{conic2022}, Lizard~\cite{graham2021lizard}, NuCLS~\cite{amgad2021nucls}, and MoNuSAC~\cite{verma2021monusac} with CONCH~\cite{lu2023conch} and DINOv3-B~\cite{simeoni2025dinov3} in the component study. The stride study uses the first three datasets with both encoders. Mask and cost studies use Lizard with CONCH, and pair averages are unweighted by cell count.

\paragraph{Specialized cell models.}
We compare with the cell-specific encoders VOLTA~\cite{nakhli2024volta} and LEMON~\cite{chadoutaud2026lemon} on the same cells and fixed partitions, using the same standardized class-balanced linear probe and metrics.

\begin{table}[!t]
\centering
\caption{\textbf{Comparison with specialized cell models.} BAcc (\%) under the matched linear-probe protocol; blue shading marks SRDP readouts.}
\label{tab:single_cell_models}
\footnotesize
\setlength{\tabcolsep}{2.8pt}
\renewcommand{\arraystretch}{1.04}
\begin{tabular}{@{}lrrrr@{}}
\toprule
& \multicolumn{4}{c}{BAcc (\%)} \\
\cmidrule(lr){2-5}
Method & \shortstack{CoNIC\\\cite{conic2022}} & \shortstack{Lizard\\\cite{graham2021lizard}} & \shortstack{NuCLS\\\cite{amgad2021nucls}} & \shortstack{MoNuSAC\\\cite{verma2021monusac}} \\
\midrule
VOLTA~\cite{nakhli2024volta} & 68.18 & 67.92 & 37.50 & 50.17 \\
LEMON~\cite{chadoutaud2026lemon} & 77.30 & 74.87 & 32.32 & 50.32 \\
\midrule
\rowcolor{methodshade}\method{} (DINOv3-B~\cite{simeoni2025dinov3}) & 71.96 & 70.50 & 38.77 & \textbf{50.66} \\
\rowcolor{methodshade}\method{} (DINOv3-L~\cite{simeoni2025dinov3}) & 71.66 & 70.71 & 38.37 & 50.53 \\
\rowcolor{methodshade}\method{} (CONCH~\cite{lu2023conch}) & 82.46 & \textbf{82.82} & 41.30 & 50.05 \\
\rowcolor{methodshade}\method{} (CONCH v1.5~\cite{ding2025titan}) & 81.11 & 80.05 & 38.37 & 49.87 \\
\rowcolor{methodshade}\method{} (GigaPath~\cite{xu2024gigapath}) & \textbf{83.69} & 81.46 & 36.70 & 50.28 \\
\rowcolor{methodshade}\method{} (UNI~\cite{chen2024uni}) & 83.01 & 82.39 & \textbf{52.05} & 49.41 \\
\bottomrule

\end{tabular}
\end{table}

Table~\ref{tab:single_cell_models} shows that SRDP is stronger than the specialized cell encoders on CoNIC~\cite{conic2022}, Lizard~\cite{graham2021lizard}, and NuCLS~\cite{amgad2021nucls}, while remaining competitive on MoNuSAC~\cite{verma2021monusac}. The advantage is largest when cell identity is expressed by nuclear appearance and contour morphology, because dense mask alignment exposes precisely these cues from the frozen backbone. The encoder-dependent spread on NuCLS and the narrow spread among MoNuSAC readouts show that SRDP reorganizes retained cell evidence rather than creating evidence absent from the pretrained features.

\paragraph{Unified cumulative readout.}
Figs.~\hyperref[fig:ablation]{\ref*{fig:ablation}(a)} and~\hyperref[fig:ablation]{\ref*{fig:ablation}(b)} add one source of spatial support at a time. Overlap pooling raises mean BAcc by $5.70$ points; the $r=2$ and $r=8$ rims add $0.45$ and $0.03$ points. The ordering mirrors the design: translated overlap sampling first removes the coarse grid bottleneck for small and boundary-crossing cells, then the compact rim adds contour morphology beyond the whole-cell mean. The broader rim overlaps evidence already represented by the other pools. Together, the rims add $0.48$ points and improve five of eight pairs, establishing dense alignment as the primary mechanism and rim structure as a phenotype-dependent refinement.

We also match descriptor width by repeating the standardized overlap vector three times and setting $C=\frac{1}{3}$, which reproduces the single-copy $C=1$ logits. We improve mean BAcc by $0.59$ points over this $3D$ control. Distinct interior and rim supports, rather than duplicated dimensions, therefore supply the added information. Grouped bootstrap intervals and the exact width-matched control are reported in supp. Tables~\ref{tab:rim-bootstrap} and~\ref{tab:matched-control}.

\paragraph{Overlap density.}
Holding the readout and probe fixed, mean BAcc rises from $56.37$ at stride $p$ to $64.64$ at $\frac{p}{2}$, then reaches $65.33$ at $\frac{p}{4}$ despite nearly quadrupling the token count again. The first $8.27$-point gain arises because half-stride sampling gives small and boundary-crossing cells multiple translated observations; further densification largely resamples accessible morphology and adds $0.69$ points. In Fig.~\hyperref[fig:ablation]{\ref*{fig:ablation}(d)}, five pairs improve slightly, Lizard with CONCH changes by $0.14$ points, and NuCLS with CONCH decreases by $1.23$ points. We identify spatial alignment, rather than token count alone, as the main source of gain and use the half-stride default.

\paragraph{Mask geometry.}
We separate contour precision from spatial registration on the full Lizard~\cite{graham2021lizard} split with CONCH~\cite{lu2023conch} by comparing the released polygon, its tight bounding box, and the polygon shifted by $(4,4)$ input pixels. All three reuse the same frozen feature map and paired evaluation examples.

\begin{table}[!ht]
\centering
\caption{\textbf{Mask geometry (\%).} A tight box remains competitive, whereas a four-pixel shift is more damaging on Lizard~\cite{graham2021lizard} with CONCH~\cite{lu2023conch}. Shading marks the default.}
\label{tab:mask_perturb}
\small
\setlength{\tabcolsep}{4.5pt}
\renewcommand{\arraystretch}{1.08}
\begin{tabular}{@{}lccc@{}}
\toprule
  Mask input & BAcc (\%) & mF1 (\%) & Acc (\%) \\
\midrule
\cellcolor{methodshade}Original polygon & \cellcolor{methodshade}\textbf{82.82} & \cellcolor{methodshade}\textbf{71.64} & \cellcolor{methodshade}86.62 \\
Tight bounding box & 82.13 & 71.48 & \textbf{86.65} \\
Shifted polygon $(4,4)$ px & 78.96 & 67.01 & 84.01 \\
\bottomrule

\end{tabular}
\end{table}

We find that the tight box remains close to the polygon in Table~\ref{tab:mask_perturb}, reducing BAcc by $0.70$ points, whereas the small translation causes a larger $3.86$-point reduction. This asymmetry supports the alignment premise of SRDP: soft coverage remains stable under a contour approximation, while accurate registration keeps the interior and rim supports on the target cell. The box preserves the cell centre and its nearby morphology, whereas translation moves coverage toward neighbouring tissue and alters both spatial pools.

\paragraph{Extraction cost.}
Finally, we test whether the overhead comes from rim pooling or dense feature extraction. We profile the frozen CONCH~\cite{lu2023conch} forward pass and fixed readout on four Lizard~\cite{graham2021lizard} crops per batch using one RTX~3090. Five warm-up iterations precede 20 measured iterations. Timing excludes data loading and one-time setup; every row has zero trainable representation parameters.

\begin{table}[!ht]
\centering
\caption{\textbf{Frozen extraction cost on Lizard~\cite{graham2021lizard} with CONCH~\cite{lu2023conch}.} Dense extraction dominates runtime and memory; RTX~3090, batch four.}
\label{tab:cost}
\small
\setlength{\tabcolsep}{2.0pt}
\renewcommand{\arraystretch}{1.08}
\begin{tabular}{@{}lrrrr@{}}
\toprule
  Readout & Tokens (\#) & Dim. (\#) & \shortstack{Latency\\(ms per cell)} & \shortstack{Peak mem.\\(MiB)} \\
\midrule
Token mean ($p$) & 196 & $D$ & 2.70 & 1548 \\
Overlap mean ($\frac{p}{2}$) & 729 & $D$ & 11.07 & 3170 \\
\rowcolor{methodshade}\method{} ($\frac{p}{2}$) & 729 & $3D$ & 11.26 & 3179 \\
\rowcolor{methodshade}\method{} ($\frac{p}{4}$) & 2809 & $3D$ & 68.53 & 3426 \\
\bottomrule

\end{tabular}
\end{table}

We localize the overhead in Table~\ref{tab:cost}. At stride $\frac{p}{2}$, both rim pools add $0.19$ ms per cell over overlap mean, whereas stride $\frac{p}{4}$ is $6.09\times$ slower than half-stride SRDP for a $0.69$-point mean BAcc gain. Dense frozen grid extraction dominates cost and supports the half-stride default.

\subsection{Discussion}
Our experiments trace one mechanism across complementary tests. Translated overlap sampling adds $5.70$ mean BAcc points, whereas the rims add $0.48$ points beyond the whole-cell pool. Quarter-stride sampling adds only $0.69$ points while nearly quadrupling token count again. Thus, patch-phase coverage removes the main bottleneck, and contour-aware pooling provides a smaller refinement. The width-matched control assigns that refinement to distinct interior and rim supports rather than dimensionality.

We isolate alignment through the mask study: a tight box changes BAcc by $-0.70$ points, whereas a four-pixel translation changes it by $-3.86$ points. Soft coverage therefore tolerates contour uncertainty but depends on registration, because translation moves both pools toward neighbouring tissue. Boundary imprecision preserves interior evidence, whereas translation redirects both supports to adjacent nuclei. This response matches our goal of assigning frozen features to the target cell.

Our downstream tasks show the same organization. On CoNIC~\cite{conic2022} and Lizard~\cite{graham2021lizard}, all six encoders improve under linear probing, and our readout remains ahead from one to 500 shots. At 500 shots, our margins remain $11.73$ and $11.05$ points. Retrieval gains are larger for macro averages, while the anchor analysis raises CONCH neutrophil AP from $0.481$ with token mean and $0.503$ with slide pooling to $0.756$ with SRDP, concentrating similarity on matching lobulated cells. Our classification and retrieval results expose complementary effects of the same descriptor. A linear probe improves when cell-centred cues become separable, while direct retrieval improves only if those cues also reorganize neighbourhoods before fitting. The persistent few-shot margin confirms that descriptor geometry remains decisive as the label budget grows.

Across datasets, gains follow the spatial evidence available to the readout. CoNIC and Lizard provide the clearest improvements, NuCLS~\cite{amgad2021nucls} varies by encoder, and the four-label MoNuSAC~\cite{verma2021monusac} evaluation is nearly tied. This pattern supports our premise: dense mask alignment exposes local evidence retained by a frozen backbone, with larger gains when that evidence matches the label semantics.

\section{Conclusion}
We introduced SRDP to treat fine-grained cell representation as a geometry-aware readout problem over frozen pathology foundation models. Our dense grid resolves patch-phase ambiguity, soft coverage registers irregular masks, and inner rims retain contour morphology. Linear-probe, few-shot, retrieval, and ablation results consistently show that these components unlock cell-scale evidence mixed by native patch pooling. We thereby connect broad pathology pretraining to cell-centred analysis through a training-free readout that transfers across backbones and tasks. By separating representation learning from spatial assignment, SRDP lets one frozen encoder serve cell typing and direct retrieval under the same geometry. This establishes readout design as a practical axis for adapting PFMs to cellular analysis. Fixed pixel-space rims require scale normalization.

\section*{LLM Usage Statement}
OpenAI Codex was used for manuscript drafting, language and \LaTeX{} editing, figure and caption refinement, code development and inspection, bibliography verification, compilation, and consistency checks. It proposed textual and code revisions. The authors determined the scientific design and experiments, evaluated every suggestion, and verified all measurements, analyses, figures, citations, claims, code, and final text.

{\small
\bibliographystyle{ieeenat_fullname}
\bibliography{references}
}

\clearpage
\setcounter{section}{0}
\renewcommand{\thesection}{S\arabic{section}}
\renewcommand{\theHsection}{S\arabic{section}}
\setcounter{table}{0}
\renewcommand{\thetable}{S\arabic{table}}
\renewcommand{\theHtable}{S\arabic{table}}
\setcounter{figure}{0}
\renewcommand{\thefigure}{S\arabic{figure}}
\renewcommand{\theHfigure}{S\arabic{figure}}
\setcounter{equation}{0}
\renewcommand{\theequation}{S\arabic{equation}}
\newif\ifstandalonesupp
\standalonesuppfalse
\section*{Supplementary Material}

\section{Data Partitions and Paired Evaluation}
\label{sec:supp:repro}
We use the same cell identifiers, labels, and split assignments for all readouts within each dataset and encoder pair. Table~\ref{tab:splits} summarizes the dataset partitions. CoNIC~\cite{conic2022} and Lizard~\cite{graham2021lizard} use fixed image partitions generated with random seed 0. NuCLS~\cite{amgad2021nucls} follows official fold~1 with no patient overlap. MoNuSAC~\cite{verma2021monusac} follows its official image split; two TCGA patients occur in both partitions, but the images are disjoint.

\begin{table}[!htbp]
  \centering
  \caption{\textbf{Dataset partitions.} Counts are numbers of units. Patient provenance is unavailable for CoNIC and Lizard.}
  \label{tab:splits}
  \footnotesize
  \setlength{\tabcolsep}{2.2pt}
  \begin{tabular}{@{}lp{0.30\linewidth}cc@{}}
    \toprule
    Dataset & Partition rule & Training units (\#) & Test units (\#) \\
    \midrule
    CoNIC~\cite{conic2022} & Image split, seed 0 & 3,985 images & 996 images \\
Lizard~\cite{graham2021lizard} & Image split, seed 0 & 190 images & 48 images \\
NuCLS~\cite{amgad2021nucls} & Official fold 1 & \shortstack{1,481 images\\102 patients} & \shortstack{263 images\\22 patients} \\
MoNuSAC~\cite{verma2021monusac} & Official split & \shortstack{209 images\\46 patients} & \shortstack{101 images\\25 patients} \\
\bottomrule

  \end{tabular}
\end{table}

Table~\ref{tab:coverage} reports the cells shared by token mean, slide pooling, and \method{} for each encoder. Feature availability varies across encoders, so encoder comparisons can involve different sample counts; readout comparisons within each dataset and encoder pair use the same cells. MoNuSAC contains 43,297 paired cells. Its test split also contains 2,322 nuclei with labels absent from the training split. We exclude these nuclei from all readouts and evaluate the four labels shared by the training and test splits. The MoNuSAC results thus use one common four-label task.

\begin{table}[!htbp]
  \centering
  \caption{\textbf{Paired feature coverage (cells).} Test counts exclude labels absent from the training split; the last column gives the number excluded. Readouts within each dataset and encoder pair use the same cells.}
  \label{tab:coverage}
  \scriptsize
  \setlength{\tabcolsep}{1.0pt}
  \renewcommand{\arraystretch}{0.90}
  \begin{tabular}{@{}llrrrr@{}}
    \toprule
    Dataset & Encoder & Paired (\#) & Train (\#) & Test (\#) & Excl. (\#) \\
    \midrule
    \multirow{6}{*}{CoNIC~\cite{conic2022}}
& DINOv3-B~\cite{simeoni2025dinov3} & 466,577 & 371,612 & 94,965 & 0 \\
& DINOv3-L~\cite{simeoni2025dinov3} & 466,577 & 371,612 & 94,965 & 0 \\
& CONCH~\cite{lu2023conch} & 466,577 & 371,612 & 94,965 & 0 \\
& CONCH v1.5~\cite{ding2025titan} & 465,757 & 371,189 & 94,568 & 0 \\
& Prov-GigaPath~\cite{xu2024gigapath} & 466,577 & 371,612 & 94,965 & 0 \\
& UNI~\cite{chen2024uni} & 359,884 & 285,911 & 73,973 & 0 \\
\midrule
\multirow{6}{*}{Lizard~\cite{graham2021lizard}}
& DINOv3-B~\cite{simeoni2025dinov3} & 361,897 & 297,345 & 64,552 & 0 \\
& DINOv3-L~\cite{simeoni2025dinov3} & 361,897 & 297,345 & 64,552 & 0 \\
& CONCH~\cite{lu2023conch} & 361,897 & 297,345 & 64,552 & 0 \\
& CONCH v1.5~\cite{ding2025titan} & 361,897 & 297,345 & 64,552 & 0 \\
& Prov-GigaPath~\cite{xu2024gigapath} & 361,897 & 297,345 & 64,552 & 0 \\
& UNI~\cite{chen2024uni} & 361,897 & 297,345 & 64,552 & 0 \\
\midrule
\multirow{6}{*}{NuCLS~\cite{amgad2021nucls}}
& DINOv3-B~\cite{simeoni2025dinov3} & 59,367 & 52,229 & 7,138 & 0 \\
& DINOv3-L~\cite{simeoni2025dinov3} & 59,367 & 52,229 & 7,138 & 0 \\
& CONCH~\cite{lu2023conch} & 59,367 & 52,229 & 7,138 & 0 \\
& CONCH v1.5~\cite{ding2025titan} & 59,349 & 52,211 & 7,138 & 0 \\
& Prov-GigaPath~\cite{xu2024gigapath} & 59,367 & 52,229 & 7,138 & 0 \\
& UNI~\cite{chen2024uni} & 47,131 & 42,899 & 4,232 & 0 \\
\midrule
\multirow{6}{*}{MoNuSAC~\cite{verma2021monusac}}
& DINOv3-B~\cite{simeoni2025dinov3} & 43,297 & 27,291 & 13,684 & 2,322 \\
& DINOv3-L~\cite{simeoni2025dinov3} & 43,297 & 27,291 & 13,684 & 2,322 \\
& CONCH~\cite{lu2023conch} & 43,297 & 27,291 & 13,684 & 2,322 \\
& CONCH v1.5~\cite{ding2025titan} & 43,297 & 27,291 & 13,684 & 2,322 \\
& Prov-GigaPath~\cite{xu2024gigapath} & 43,297 & 27,291 & 13,684 & 2,322 \\
& UNI~\cite{chen2024uni} & 43,297 & 27,291 & 13,684 & 2,322 \\
\bottomrule

  \end{tabular}
\end{table}

We standardize features using the training cells and fit class-balanced multinomial logistic regression with an $\ell_2$ penalty, inverse regularization strength $C=1$, and limited-memory BFGS optimization. Retrieval uses the complete training split as the gallery and the test split as queries. For macro recall at rank $K$ (mR@$K$), we average the hit indicator within each represented query class and then weight classes uniformly. All probe and retrieval evaluations use fixed splits. For each shot budget, we draw ten class-balanced subsets with random seed 0 and keep the test set fixed.

\section{Class-Balanced Retrieval Beyond Rank One}
\label{sec:supp:retrieval}
The main paper reports R@1 and mR@1. Table~\ref{tab:macro-retrieval} extends the class-balanced view to ranks 5 and 10. On CoNIC~\cite{conic2022}, \method{} improves mR@5 by 6.70 points and mR@10 by 4.68 points over token mean. On Lizard~\cite{graham2021lizard}, the corresponding gains are 8.36 and 5.86 points. The gains at both ranks indicate improved local same-class neighbourhoods in the frozen descriptor space.

NuCLS~\cite{amgad2021nucls} changes by 0.25 points at rank 5 and by $-0.14$ points at rank 10, while MoNuSAC~\cite{verma2021monusac} changes by $-0.50$ and $-0.86$ points. These smaller changes mirror the main linear-probe results. Across tasks, the retrieval gains track how strongly the frozen lattice preserves label-relevant morphology at cell scale.

\begin{table}[!htbp]
  \centering
  \caption{\textbf{Additional class-balanced retrieval (\%).} Results first average query classes uniformly and then average the six encoders. Higher is better.}
  \label{tab:macro-retrieval}
  \ifstandalonesupp
    \scriptsize
    \renewcommand{\arraystretch}{0.92}
  \else
    \footnotesize
  \fi
  \setlength{\tabcolsep}{5.0pt}
  \begin{tabular}{@{}llcc@{}}
    \toprule
    Dataset & Readout & mR@5 & mR@10 \\
    \midrule
    \multirow{3}{*}{CoNIC~\cite{conic2022}} & Token mean & 76.79 & 84.20 \\
& Slide pooling & 79.59 & 86.25 \\
\rowcolor{methodshade}& \method{} (ours) & \textbf{83.49} & \textbf{88.88} \\
\midrule
\multirow{3}{*}{Lizard~\cite{graham2021lizard}} & Token mean & 72.03 & 80.40 \\
& Slide pooling & 75.24 & 83.20 \\
\rowcolor{methodshade}& \method{} (ours) & \textbf{80.39} & \textbf{86.26} \\
\midrule
\multirow{3}{*}{NuCLS~\cite{amgad2021nucls}} & Token mean & 58.25 & \textbf{67.63} \\
& Slide pooling & 56.37 & 66.61 \\
\rowcolor{methodshade}& \method{} (ours) & \textbf{58.50} & 67.49 \\
\midrule
\multirow{3}{*}{MoNuSAC~\cite{verma2021monusac}} & Token mean & 56.37 & 59.76 \\
& Slide pooling & \textbf{57.61} & \textbf{61.90} \\
\rowcolor{methodshade}& \method{} (ours) & 55.87 & 58.90 \\
\bottomrule

  \end{tabular}
\end{table}

\section{Radius Selection and Statistical Controls}
\label{sec:supp:radius}
\paragraph{Training-only radius selection.}
We select the rim radii on a grouped validation split constructed from each training partition. Fold~0 of a five-fold \texttt{StratifiedGroupKFold} uses seed 2027. Images define groups for CoNIC~\cite{conic2022} and Lizard~\cite{graham2021lizard}; patients define groups for NuCLS~\cite{amgad2021nucls} and MoNuSAC~\cite{verma2021monusac}. Candidate pairs are ranked by unweighted validation BAcc across the eight dataset and encoder pairs used in the cumulative ablation. The selected pair is $(2,8)$ (Table~\ref{tab:radius-selection}); the two leading pairs, $(2,8)$ and $(2,4)$, differ by 0.04 BAcc points.

\begin{table}[!htbp]
  \centering
  \caption{\textbf{Training-only radius selection.} Mean validation BAcc is an unweighted average over eight pairs.}
  \label{tab:radius-selection}
  \ifstandalonesupp
    \scriptsize
    \renewcommand{\arraystretch}{0.92}
  \else
    \small
  \fi
  \setlength{\tabcolsep}{6pt}
  \begin{tabular}{lcc}
    \toprule
    Candidate radii (px) & Mean BAcc (\%) & Selected \\
    \midrule
    $(2,4)$ & 70.61 & $\times$ \\
    $(2,8)$ & \textbf{70.65} & $\checkmark$ \\
    $(4,8)$ & 69.98 & $\times$ \\
    \bottomrule
  \end{tabular}
\end{table}

\paragraph{Grouped bootstrap.}
Each of 5,000 bootstrap draws resamples the highest available grouping unit, retaining all cells from each sampled image for CoNIC and Lizard or patient for NuCLS and MoNuSAC. We compute paired BAcc differences within each dataset and encoder pair and then average the eight pairs uniformly. The narrow rim contributes $+0.45$ points with a 95\% confidence interval of $[0.11,1.07]$, whereas the wider rim contributes an additional $+0.03$ points with an interval of $[-0.29,0.19]$ (Table~\ref{tab:rim-bootstrap}). Most of the observed rim gain is associated with the compact contour band.

\begin{table}[!htbp]
  \centering
\caption{\textbf{Grouped bootstrap rim analysis.} BAcc differences and 95\% confidence intervals are in percentage points; the final column counts positive dataset and encoder pairs.}
  \label{tab:rim-bootstrap}
  \ifstandalonesupp
    \scriptsize
    \renewcommand{\arraystretch}{0.90}
  \else
    \footnotesize
  \fi
  \setlength{\tabcolsep}{0.6pt}
  \begin{tabular}{@{}p{0.40\linewidth}ccc@{}}
    \toprule
    Comparison & \shortstack{$\Delta$BAcc\\(pp)} & \shortstack{95\% CI\\(pp)} & \shortstack{Positive\\pairs (\#)} \\
    \midrule
    Overlap plus $r=2$ vs. overlap & $+0.45$ & $[0.11,1.07]$ & 4 of 8 \\
    Complete \method{} vs. overlap plus $r=2$ & $+0.03$ & $[-0.29,0.19]$ & 5 of 8 \\
    Complete \method{} vs. overlap & $+0.48$ & $[0.10,1.01]$ & 5 of 8 \\
    \method{} at $C=\frac{1}{3}$ vs. matched repeated overlap & $+0.59$ & $[0.24,1.19]$ & 4 of 8 \\
    \bottomrule
  \end{tabular}
\end{table}

\paragraph{Width-matched control.}
Let $x$ be the standardized overlap descriptor and $\beta$ the optimum coefficient vector at $C=1$. Repeating the descriptor as $[x,x,x]$ and assigning $\beta/3$ to each block preserves every logit. The squared coefficient norm becomes one third as large, so $C=1/3$ restores the original $\ell_2$ penalty. The repeated-overlap model matches the predictions and probabilities of the single-copy overlap model. Under the same regularization, \method{} increases mean BAcc from 60.59 to 61.17 (Table~\ref{tab:matched-control}). This comparison separates the contribution of distinct interior and rim supports from descriptor width.

\begin{table}[!htbp]
  \centering
  \caption{\textbf{Regularization-matched control.} $C$ is inverse $\ell_2$ regularization strength. Metrics are unweighted means over eight pairs.}
  \label{tab:matched-control}
  \ifstandalonesupp
    \scriptsize
    \renewcommand{\arraystretch}{0.92}
  \else
    \footnotesize
  \fi
  \setlength{\tabcolsep}{3.0pt}
  \begin{tabular}{lccccc}
    \toprule
    Readout & Dim. (\#) & $C$ & BAcc (\%) & mF1 (\%) & Acc (\%) \\
    \midrule
    Overlap & $D$ & 1 & 60.59 & 53.33 & 78.40 \\
    Repeated overlap & $3D$ & $\frac{1}{3}$ & 60.59 & 53.33 & 78.40 \\
    \method{} & $3D$ & $\frac{1}{3}$ & \textbf{61.17} & \textbf{54.55} & \textbf{79.31} \\
    \method{} (default) & $3D$ & 1 & 61.07 & \textbf{54.57} & 79.25 \\
    \bottomrule
  \end{tabular}
\end{table}
\FloatBarrier

The cumulative ablation uses all four datasets with CONCH~\cite{lu2023conch} and DINOv3-B~\cite{simeoni2025dinov3}. The stride study uses the six dataset and encoder pairs from CoNIC, Lizard, and NuCLS for which all three stride settings were evaluated. Mask geometry and cost are evaluated on Lizard with CONCH; cost is measured on one RTX~3090. Pair-level summaries use an unweighted mean across dataset and encoder pairs. All comparisons use the paired cells, probe settings, and labels described in Sec.~\ref{sec:supp:repro}.

\end{document}